\documentclass{sig-alternate-05-2015}
\usepackage{algorithm}
\usepackage{algpseudocode}
\usepackage{pifont}
\usepackage{multirow}
\usepackage{graphicx}
\usepackage{subfigure}

\begin{document}

\setcopyright{acmcopyright}

%
\CopyrightYear{2016}
\setcopyright{acmcopyright}
\conferenceinfo{MM '16,}{October 15-19, 2016, Amsterdam, Netherlands}
\isbn{978-1-4503-3603-1/16/10}\acmPrice{\$15.00}
\doi{http://dx.doi.org/10.1145/2964284.2967273}


\title{Scalable Compression of Deep Neural Networks}
%
%
%
%
%
%
\numberofauthors{2} 
%
\author{
%
%
\alignauthor
Xing Wang \\
       \affaddr{Simon Fraser University, BC, Canada}\\
       \affaddr{AltumView Systems Inc., BC, Canada}\\
       \email{xingw@sfu.ca}
\alignauthor
Jie Liang\\
       \affaddr{Simon Fraser University, BC, Canada}\\
       \affaddr{AltumView Systems Inc., BC, Canada}\\
       \email{jiel@sfu.ca}
}

 \ccsdesc[500]{Computing methodologies~Neural networks}
 \ccsdesc[100]{Information systems~Clustering and classification}
\ccsdesc[300]{Computing methodologies~Discrete space search}
\maketitle
\begin{abstract}
Deep neural networks generally involve some layers with millions of parameters, making them difficult to be deployed and updated on devices with limited resources such as mobile phones and other smart embedded systems. In this paper, we propose a scalable representation of the network parameters, so that different applications can select the most suitable bit rate of the network based on their own storage constraints. Moreover, when a device needs to upgrade to a high-rate network, the existing low-rate network can be reused, and only some incremental data are needed to be downloaded. We first hierarchically quantize the weights of a pre-trained deep neural network to enforce weight sharing. Next, we adaptively select the bits assigned to each layer given the total bit budget. After that, we retrain the network to fine-tune the quantized centroids. Experimental results show that our method can achieve scalable compression with graceful degradation in the performance.
\end{abstract}

%
%

%
%

%
%
\printccsdesc


\keywords{Deep neural network, scalable compression}

\section{Introduction}

Deep neural networks (DNN) or deep learning have evolved into the state-of-the-art technique for many artificial intelligence tasks including computer vision \cite{AlexNet}, \cite{VGG}. In this paper, we focus on the convolutional neural network (CNN), which was originally developed in 1998 by LeCun et al. with less than 1M parameters (weights) to classify handwritten digits \cite{CNN}. In 2012, CNN was used as a key component in \cite{AlexNet} to achieve the breakthrough in ImageNet Large Scale Visual Recognition Challenge 2012 (ILSVRC2012), and the proposed AlexNet has 60M parameters and needs 240MB of storage space. In 2014, Simonyan et al. further improved the accuracy by $10\% $  \cite{VGG}, and the VGG-16 model they developed has 138M parameters.

Although the performance of DNN is very promising, its application in low-end devices such as mobile phones faces some challenges. For example, many devices have limited storage spaces. Therefore storing millions of DNN parameters on these devices could be a problem. If the DNN network needs to be updated, usually via wireless channels, downloading the large amount of network parameters will cause excessive delay. Moreover, running large-scale DNNs with floating-point parameters could consume too much energy and slow down the algorithm. Therefore, efficient compression of the DNN parameters without sacrificing too much the performance becomes an important topic.

There have been some recent works on the compression of neural networks. Vanhoucke et al. \cite{FixedQt} proposed a fixed-point implementation with 8-bit integer (vs 32-bit floating-point) activations. Denton et al. \cite{NIPS2014_5544} exploited the linear structure of the neural network by finding an appropriate low-rank approximation of the parameters and keeping the accuracy within $1\%$ of the original model. Kim et al. \cite{ICLR16_Mobile} applied tensor decomposition to the network parameters and proposed an one-shot whole network compression scheme that can achieve significant reductions in model size, runtime and energy consumption.

Many works focus on binning the network parameters into buckets, and only the values in the bucket need to be stored. HashedNets \cite{icml2015_chenc15} is a recent technique to reduce model size by using a hash function to randomly group connection weights, so that all connections within the same hash bucket share a single parameter value. Gong et al. \cite{VecQt} compressed deep convnets using vector quantization, which resulted in $1\%$ accuracy loss. Both methods studied the fully-connected (FC) layer in the CNN, but ignored the convolutional (CONV) layers. Recently, Han et al. \cite{ICLR16} introduced a deep neural network compression pipeline by combining pruning, quantization and Huffman encoding, which can reduce the storage requirement of neural network by 35 $ \times $ or 49 $ \times $ without affecting their accuracy.

In this paper, motivated by the successful applications of scalable coding in various image and video coding standards such as JPEG 2000, H.264, and H.265/HEVC \cite{J2K,SVC,SHEVC}, we propose a scalable compression framework for DNNs, which has not been addressed before. Our goal is to represent the DNN parameters in a scalable fashion such that we can easily truncate the representation of the network according to the storage constraint and still get near-optimal performance at each rate. Moreover, if the network needs to be upgraded with higher rate and better performance, the existing low-rate network can be reused, and only some incremental data are needed. This is better than recompressing and re-transmitting the network as in \cite{ICLR16,ICLR16_Mobile}.

To achieve this goal, we propose a three-stage pipeline. First, a hierarchical representation of weights in DNNs is developed. Second, we propose a backward greedy search algorithm to adaptively select the bits assigned to each layer given the total bit budget. Finally, we fine-tune the compressed model.

The rest of the paper is organized as follows. Sec. \ref{Hierarchical} is devoted to hierarchical quantization of the DNN parameters. In Sec. \ref{Adaptive} we formulate the bit allocation as an optimization problem and propose a backward search solution. A fine-tuning method is presented in Sec. \ref{Pipeline}. Experimental results on MNIST, CIFAR-10 and ImageNet datasets are reported in Sec. \ref{Experiments}, followed by conclusions in Sec. \ref{sec_con}.

\section{Hierarchical Quantization}
\label{Hierarchical}

The K-means clustering-based quantization is a popular technique in the compression of DNN \cite{VecQt}, \cite{ICLR16}. Therefore, in this paper, we also choose K-means clustering with linear initialization \cite{ICLR16} to compress the weights in DNN. However, the framework developed in this paper is quite general and can also be applied to other quantization techniques, e.g., the fixed-point quantization in \cite{FixedQt} and other similar tasks besides classification, e.g., regression problems.

In \cite{ICLR16}, the authors quantize the weights to enforce weight sharing with K-means clustering, e.g., they assign 8 bits (256 shared weights) to each CONV layer and 5-bits (32 shared weights) to each FC layer. However, every time a CONV layer is assigned a different bit, the K-means clustering has to be performed again, rendering scalable compression infeasible. On the other hand, some DNN layers have a large number of weights, e.g., the number of weights in the fc6 layer of AlexNet is 38M. Therefore the K-means clustering can be quite slow, even with the help of GPU.

To address this problem, we adopt the scalable coding concept in image/video coding \cite{J2K,SVC,SHEVC}, and represent the weights hierarchically, i.e., each weight is represented by a base-layer component and several enhancement-layer components; hence, we only need to perform the quantization step once during the entire scalable compression process, which also benefits the adaptive bit allocation in Sec. \ref{Adaptive}. Note that there are two different kinds of layers in this paper: the network layers in DNN, and the hierarchical quantization layers in the scalable representation of the weights.

Suppose we want to allocate $n$ bits to each weight in a pre-trained DNN layer. We first perform K-means clustering of all weights with $K=2$ (1-bit quantization), and record the corresponding cluster indices and centroids. We also record the corresponding quantization error. This yields the 1-bit base-layer approximations of all weights. Next, we perform another K-means clustering with $K=2$ on all quantization errors, and record the corresponding cluster indices, centroids, and quantization errors. The gives us the 1-bit first-enhancement-layer representations of all weights. By repeating this procedure, we can obtain a $n$-layer hierarchical representation of a weight, i.e.,
 \begin{equation}
 w \approx {b_1} + {e_1} + ... + {e_{n - 1}},
 \end{equation}
where $w$ is a uncompressed weight, $b_1$ and $e_i$ are the centroid of the base layer and the $i$-th enhancement layer respectively.

This hierarchical quantization only needs to be performed once, which facilitates future network updating, as we only need to add or delete certain quantization layers to meet the new bit rate constraint. For the tradition K-means clustering used in \cite{VecQt} and \cite{ICLR16}, we have to perform K-means clustering every time a new bit budget is required.

After the hierarchical quantization, we can build a codebook that stores the centroid and cluster index information of all quantization layers. For a network layer of DNN with $N$ weights, there are $2n$ centroids, and the number of cluster indices is $Nn$. If each uncompressed weight or centroid is represented by $b$ bits ($b$=32 for single-precision floating-point number), the compression rate of the $n$-bit hierarchical quantization scheme is
\begin{equation}
r = \frac{{Nb}}{{Nn + 2nb}}.
\end{equation}
In contrast, in the conventional K-means method \cite{ICLR16}, given the same $n$-bit quantization, the compression rate is $r = Nb/(Nn + {2^n}b)$.

Note that the storage cost is dominated by $Nn$, compared to $2nb$ or $2^nb$, because the number of connections $N$ in a DNN is usually very large.

\section{Adaptive bit allocation}
\label{Adaptive}

In DNN, the redundancies in different network layers are different \cite{ICLR16,FixedQt}. Therefore it is necessary to design an optimal bit allocation algorithm, i.e., given a bit budget, how to allocate the bits to different network layers in order to get the best performance. In this paper, we formulate the following optimization problem.
\begin{equation}
\label{Problem1}
\begin{array}{l}
\arg \mathop {{\rm{     }} \min }\limits_{\{ \bf{n}, \bf{C}, \bf{G}\} } f(\bf{n},\bf{C},\bf{G})\\
{\rm{s}}{\rm{.t}}{\rm{.   }}\sum\limits_{i = 1}^L {{N_i}{n_i} + 2{n_i}b}  \le \mu.
\end{array}
\end{equation}
where $\bf{n} = \left[ {\begin{array}{*{20}{c}}
{{n_1}}&{...}&{{n_L}}
\end{array}} \right]$ is a vector containing the bits allocated to $L$ network layers, $\bf{C}$ is the centroid vector, $\bf{G}$ is the cluster-by-index matrix for the network layers, ${N_i}$ is the number of weights in the $i$-th network layer, and $\mu$ is the bit budget. We use the cross entropy between the pdf of the predicted labels and true labels as the cost function $f( \cdot )$, which is frequently used in classification tasks.

It is hard to solve the combinatorial optimization in Eq. (\ref{Problem1}), since the number of bits assigned to each network layer ${n_i}$ has to be integer and the number of entries in the cluster-by-index matrix $\bf{G}$ is $\sum\limits_{i = 1}^L {{N_i}{n_i}}$, even larger than the number of weights $\sum\limits_{i = 1}^L {{N_i}} $ in the pre-trained DNN. Therefore, we use a similar method to \cite{ICLR16} to first approximate the original uncompressed weights with high-rate quantized weights. More specifically, we first use the hierarchical method in Sec. \ref{Hierarchical} to assign $M$ bits to each CONV layer weight and $P$ bits to each FC layer weight. This is used as the initialization step. The centroid vector $\bf{C}$ and the cluster-by-index matrix $\bf{G}$ are then determined and fixed. We use ${\rm E}$ to denote the number of bits to store this initial network.

Next, we adaptively allocate bits to network layers such that $u < {\rm E}$. The problem in Eq. (\ref{Problem1}) is simplified to
\begin{equation}
\label{Problem2}
\begin{array}{*{20}{l}}
\quad \quad {\arg \mathop {\min }\limits_{\{ {\bf{n}}\} } f({\bf{n}})}\\
{{\rm{s}}.{\rm{t}}. \;\; {\rm{ }} B = \sum\limits_{i = 1}^L {({N_i} + 2b){n_i}}  \le \mu. {\rm{  }}}
\end{array}
\end{equation}

For small-scale problems, the optimization above can be solved by exhaustive grid search, where configurations that violate the bit constraint are skipped, and the others are evaluated to find the best solution. The process can be accelerated by parallel computing, since different configurations are independent. However, for large-scale problems, exhaustive search becomes infeasible, as the number of configurations grows exponentially with the number of bits. For example, in AlexNet, there are 5 CONV layers and 3 FC layers. If 10 bits are assigned to each CONV layer and 5 bits are assigned to each FC layer, the total number of configurations would be ${10^5} \times {5^3} = 12.5{\rm{M}}$.

One way to speed up the process is to use random search \cite{RandomSearch}, since the number of bits assigned to each network layer can be treated as a hyper-parameter for the DNN. Theoretical analysis in \cite{RandomSearch} shows that randomly selecting 60 configurations can ensure that the top $5\% $ result can be achieved with a probability of $0.95$. For the bit allocation problem here, we should randomly select a number of configurations that satisfy the bit constraint. In Sec. \ref{Experiments}, random search is used as a baseline algorithm for comparison.

In this paper, we propose a backward greedy search algorithm to address the bit constraint explicitly and solve the problem in Eq. (\ref{Problem2}). We start from the initial high-rate quantized network as discussed above. Denote the bit allocation in the $t$-th iteration as ${{\bf{n}}^t}=[n_1^t, \ldots, n_L^t]$, whose corresponding total bit cost is ${B^t}$. To find ${\bf{n}}^{t+1}$ at iteration $t+1$, we follow the spirit of the gradient descent method by assigning one less bit to each network layer respectively, calculating the corresponding gradient of the total bit cost, and choosing the configuration that has the maximum gradient. In other words, let ${{\bf{n}}^{t,j}} = [n_1^t, \ldots, n_{j-1}^t, n_j^t - 1, n_{j+1}^t, \ldots, n_L^t]$, the bit allocation in the $(t+1)$-th iteration is obtained by
\begin{equation}
\label{gradient}
\begin{array}{*{20}{l}}
{{\rm{  \qquad   }}\arg \mathop {\max }\limits_{{{\bf{n}}^{t,j}}} \frac{{f({{\bf{n}}^{t,j}}) - f({{\bf{n}}^t})}}{{{B^{t,j}} - {B^t}}}}\\
\begin{array}{l}
{\rm{s}}{\rm{.t}}{\rm{. \indent }}{{\bf{n}}^{t,j}} \subset \{ {{\bf{n}}^{t,1}},{{\bf{n}}^{t,2}},...,{{\bf{n}}^{t,L}}\} ,{\rm{ }}\\
{\rm{   \qquad   }}{B^{t,j}} = \sum\limits_{i = 1}^L {({N_i} + 2b)n_i^{t,j}}.
\end{array}
\end{array}
\end{equation}
The iteration terminates until the bit constraint is satisfied. The entire backward greedy search algorithm is summarized in Alg. \ref{backward}. The intuition behind the gradient defined above is twofold. First, if two bit allocations have the same cost function value, the one with smaller total bit cost should be chosen. Second, if two bit allocations have the same total bit cost, we should choose the one with lower cost function value and use the maximum function in Eq. (\ref{gradient}) due to ${B^{t,j}} < {B^t}$.

\begin{algorithm}
\caption{Backward Greedy Search Algorithm}
\label{backward}
\begin{algorithmic}[1]
\State \textbf{Initialization}: Quantize the network with M bits for each CONV layer and P bits for each FC layer. Let $t = 0$.
\While{${B^t} > \mu $}
\For{each network layer $j \le L$}
\State $n_j^{t,j} \leftarrow n_j^t - 1,{\rm{ }}n_p^{t,j} \leftarrow n_p^t{\rm{ \,for \,}}p \ne j$.
\State Update the weights of DNN based on the hierarchical framework in Sec. \ref{Hierarchical}
\State Test with the validation data and record ${B^{t,j}}$ and $f({{\bf{n}}^{t,j}})$
\EndFor
\State Select ${{\bf{n}}^{t + 1}}$ based on Eq. (\ref{gradient})
\State $t \leftarrow t + 1$
\EndWhile
\end{algorithmic}
\end{algorithm}
\vskip -10pt

\section{Fine Tuning (FT)}
\label{Pipeline}

It is shown in \cite{ICLR16,FixedQt} that fine-tuning (FT) of the centroids after the quantization of DNN can significantly improve the classification performance. In this paper, we also perform fine-tuning after the adaptive bit allocation to update the centroids based on Eq. (3) in \cite{ICLR16}.

The advantage of the proposed scalable compression of the DNN is that for each target bit rate, we can find a near-optimal bit allocation. If later on the DNN bit rate on a device needs to be updated, instead of re-transmitting a new set of the DNN parameters, we only need to transmit some incremental data, including the centroid vector $\bf{C}$ and cluster-by-index matrix $\bf{G}$. The required bit rate is thus much lower than replacing the entire network.

During the update, some additional bits caused by the fine-tuning are needed to update the centroids of the previous compressed model. However, according to  the analysis in Sec. \ref{Hierarchical}, the centroid update will cost $2b\sum\limits_{i = 1}^L {{n_i}} $ at most, while the minimal bits needed to update the cluster-by-index matrix are $\min \{ {N_1},{N_2},....,{N_L}\} $.
The storage cost is dominated by the cluster indices instead of centroids; hence the overhead introduced by the fine-tuning is negligible. Take AlexNet as an example, if we use 10 bits to quantize CONV layers and 5 bits for FC layers, at most 0.52KB are needed to update these centroids, while we may use at least 5KB to update the cluster-by-index matrix every time a different bit budget is given.

\section{Experimental results}
\label{Experiments}

We test the proposed scalable compression on 3 networks designed for the MNIST \cite{CNN}, CIFAR-10 \cite{Krizhevsky2009a} and ImageNet \cite{ILSVRC15} datasets respectively. We implement the network training based on the CNN toolbox MatConvNet \cite{vedaldi15matconvnet} with our own modifications. The training is done on a desktop with a \text{NVIDIA TIAN X} GPU with 12GB memory.

\subsection{LeNet-5 for MNIST}

\begin{figure}[tb*]
\begin{center}
\begin{tabular}{c@{\hspace{-0.1mm}}c}
  \includegraphics[width=1.61in]{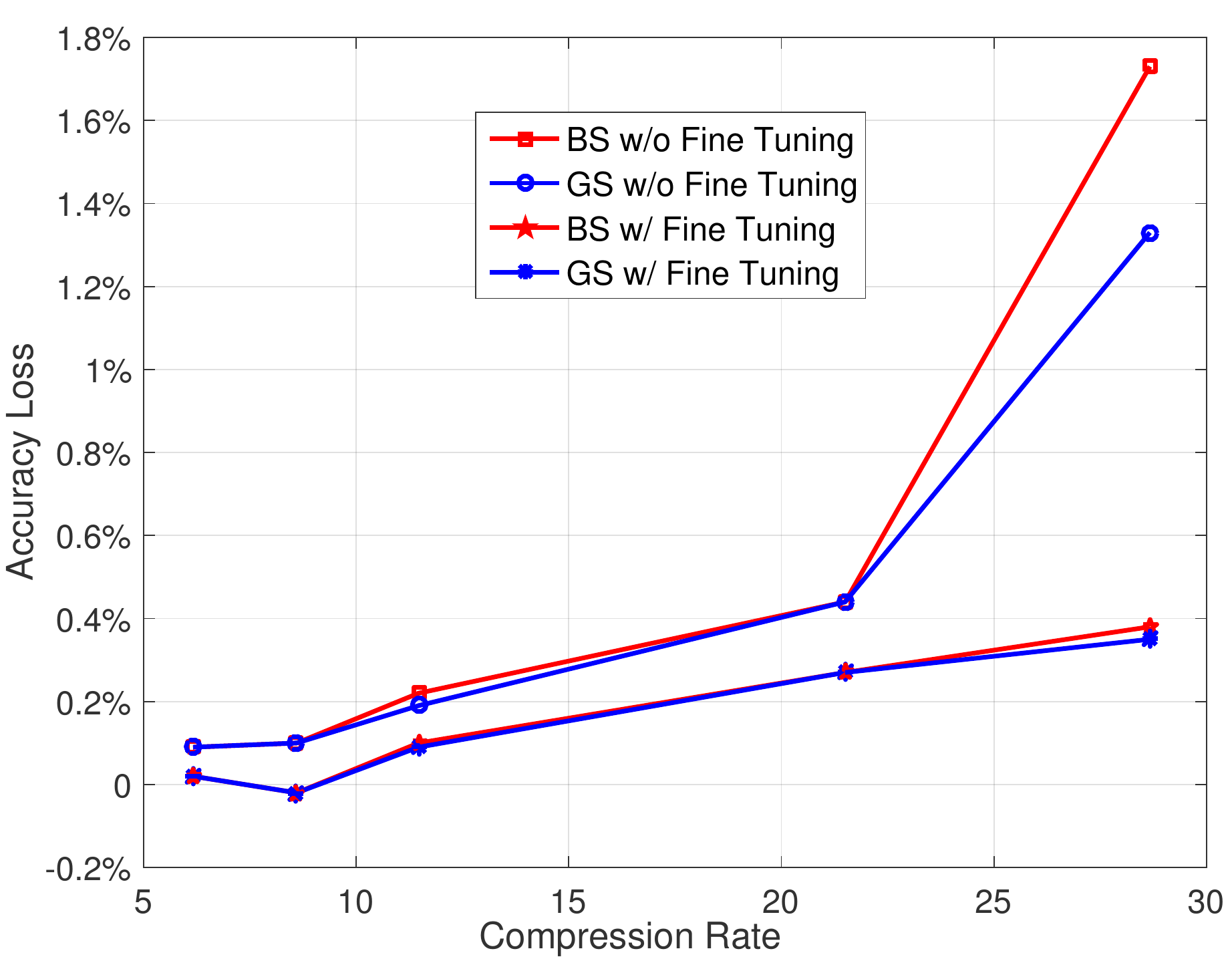} &
  \includegraphics[width=1.61in]{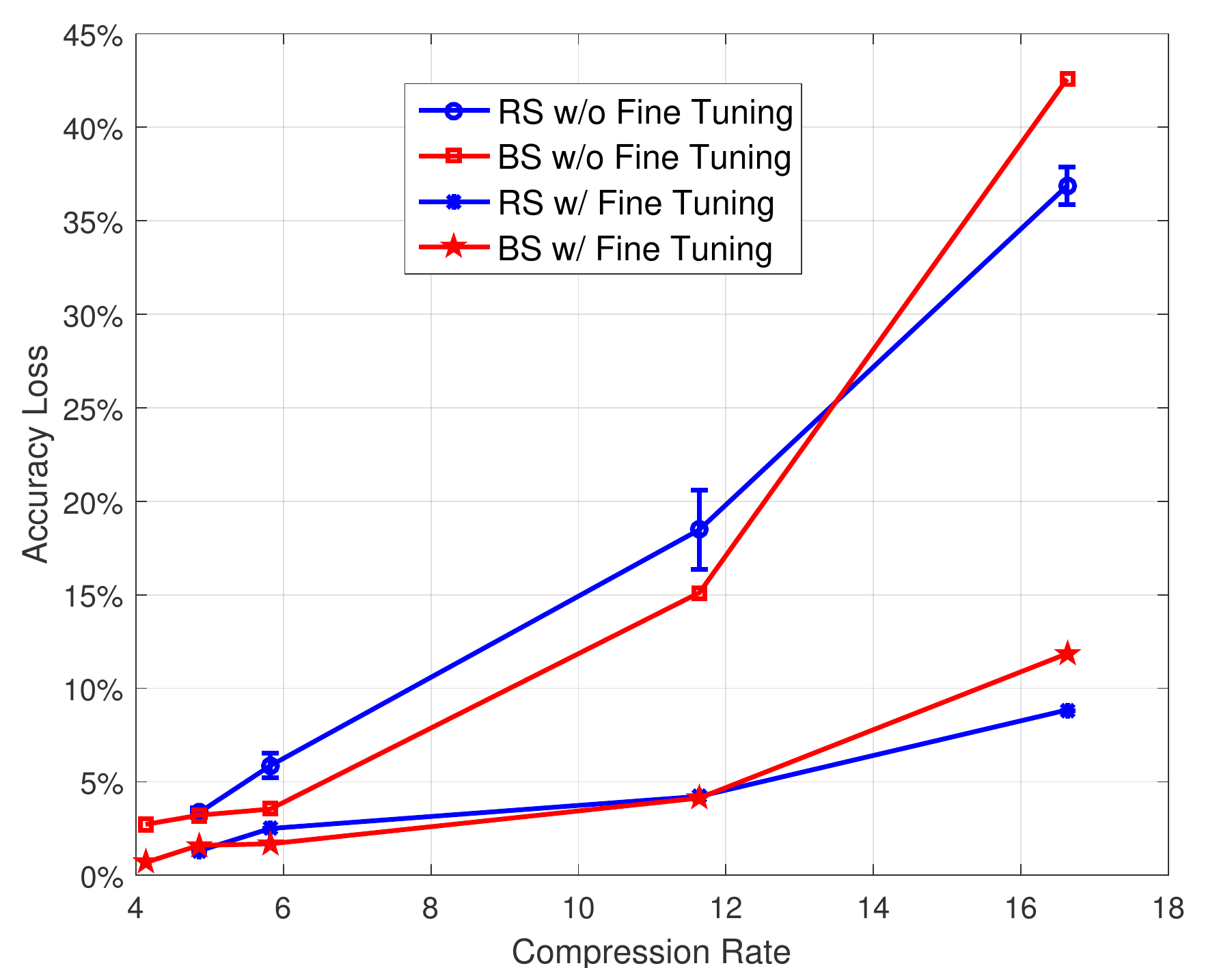} \\
  (a) & (b)
\end{tabular}
\end{center}
\vskip -15pt
\caption{\label{GridSearchMnistCifar} Top-1 accuracy loss of compressed DNNs under different bit allocation methods. (a) LeNet-5 and (b) CIFAR-10-quick. }
\vskip -5pt
\end{figure}

We use the \textit{{\rm{cnn\_mnist\_experiment}}{\rm{.m}}} function in MatConvNet to train LeNet-5 for MNIST dataset. There are 2 CONV layers and 2 FC layers. The pre-trained model can achieve 0.88$\%$ Top-1 error and needs a storage of 1720KB. We use 8 bits to hierarchically quantize each CONV layer and 5 bits for each FC layer. The initial quantized model can achieve 0.97$\%$ Top-1 error, and the corresponding storage cost is 279KB. In Fig. \ref{GridSearchMnistCifar}(a), we compare the proposed backward greedy search method (BS) with the exhaustive grid search method (GS). We also present the number of configurations tested on the validation set in Table \ref{MnistNum} to compare the computational complexity. We can see that our proposed backward search algorithm can achieve comparable compression performance to the grid search with much smaller computational complexity. The only exception happens when the compression rate is extremely large, e.g., 28.67 in Fig. \ref{GridSearchMnistCifar}(a). However, after fine tuning, the performance is still very close to the original one.


\begin{table}[tb]
\begin{center}
\begin{tabular}{c|c|c|c|c}
\hline
\multicolumn{5}{c}{LeNet-5 for MNIST} \\
\hline
Bit Budget (KB) & 200   & 150  & 80  & 60  \\
\hline
 Compression Rate & 8.60 & 11.47 & 21.50 & 28.67  \\
\hline
 BS Number & 26 & 51 & 101 & 115 \\
\hline
 GS Number & 960 & 640 & 320 & 79  \\
\hline

\multicolumn{5}{c}{CIFAR-10-quick for CIFAR-10} \\
\hline
Bit Budget (KB) & 120 & 100 & 50 & 30\\
\hline
Compression Rate  & 4.85 & 5.82  & 11.64 & 19.40 \\
\hline
 BS Number & 26 & 51  & 101 & 115  \\
\hline
RS Number &  \multicolumn{4}{c}{120} \\
\hline
\end{tabular}
\caption{\label{MnistNum}
Number of configurations tested on MNIST and CIFAR-10 validation set v.s. compression rate.}
\end{center}
\vskip -15pt
\end{table}

\subsection{CIFAR-10-quick for CIFAR-10}

We use the provided \textit{{\rm{cnn\_cifar}}{\rm{.m}}} in MatConvNet to train CIFAR-10-quick for CIFAR-10 dataset. There are 3 CONV layers and 2 FC layers in the network. The reference model can achieve $19.97\% $ Top-1 error and needs a storage space of 582KB. We use 10 bits to quantize each CONV layer and 5 bits for each FC layer. The initial quantized model can achieve $22.70\% $ Top-1 error and needs 141KB storage space. Since there are at most 25K configurations which takes too much time to evaluate, instead of using grid search as a comparison, we use the random search method (RS) \cite{RandomSearch}. In each trial, we randomly choose 120 configurations that satisfy the bit constraint from the configuration pool.

The compression performance is shown in Fig. \ref{GridSearchMnistCifar}(b) and the computational complexity is presented in Table \ref{MnistNum}. It can be seen that the proposed backward search algorithm can achieve similar or even better performance than random search with much smaller computational complexity, especially when the bit rate is close to that of the initial quantized network. The only exception happens when the compression rate is extremely large, e.g., 20 in Fig. \ref{GridSearchMnistCifar}(b).  For the fine-tuning in the random search method, we fine-tune the result that achieves the median classification accuracy in the 10 trials.



\subsection{AlexNet for ILSVRC12}

We use the provided \textit{{\rm{cnn\_imagenet}}{\rm{.m}}} to train AlexNet for ILSVRC12. The reference model is slightly different from that of the original AlexNet in \cite{AlexNet}, where the order of pooling layer and norm layer are swapped. It contains 5 CONV layers and 3 FC layers. This reference model can achieve $41.39\%$ Top-1 error, $18.85\%$ Top-5 error, and needs 240 MB to store. We use 10 bits to quantize each CONV layer and 5 bits for each FC layer. This initial quantized model can achieve $56.09\%$ Top-1 error, $31.63\%$ Top-5 error, and needs 39.5 MB to store. The number of configurations in each trial of  RS is 150. The number of trials is 5 in order to get $0.95$ confidence interval. The result that achieves the median classification accuracy in the 5 trials is fine-tuned.

\begin{figure}[tb*]
\begin{center}
\begin{tabular}{c@{\hspace{-0.1mm}}c}
  \includegraphics[width=1.61in]{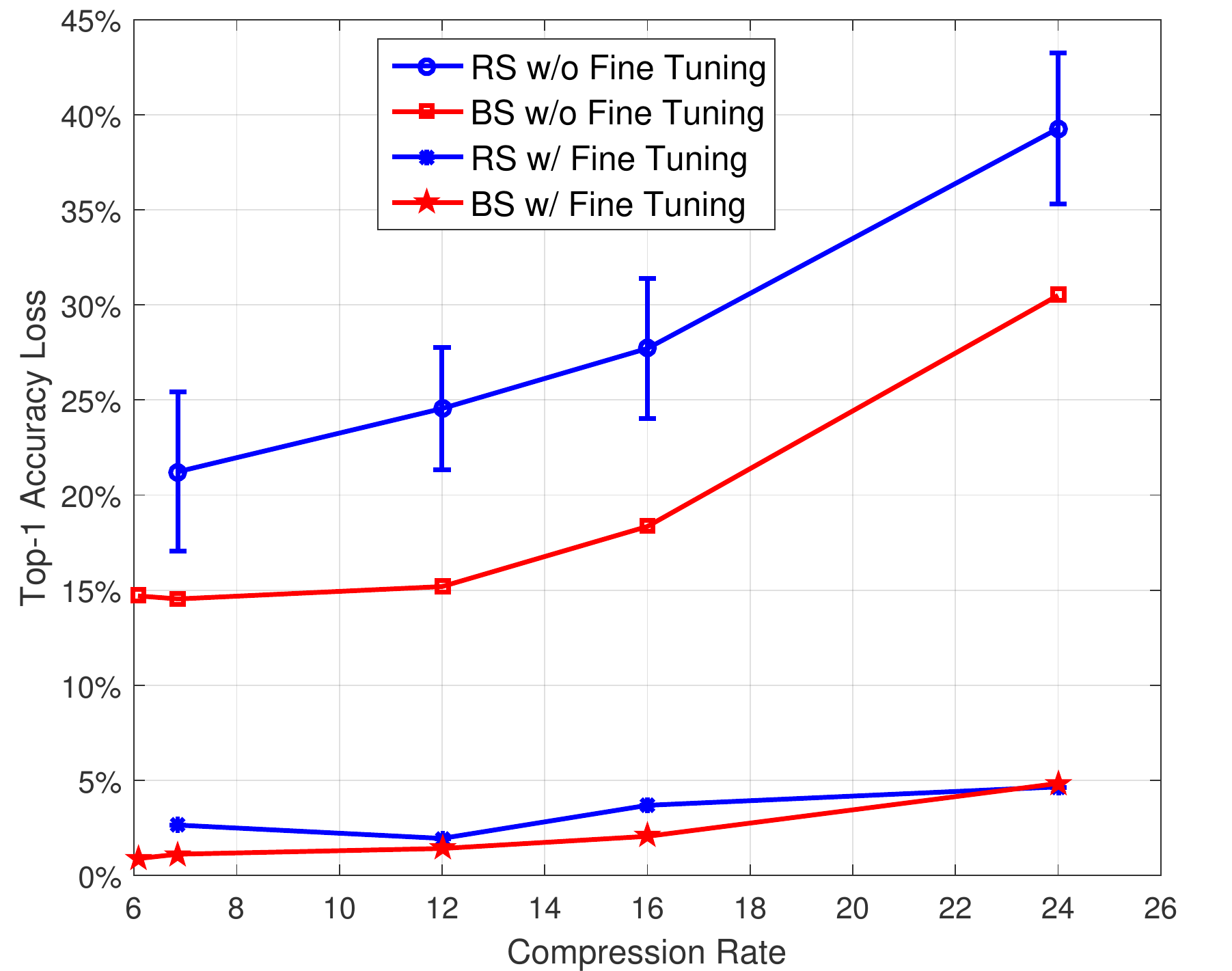} &
  \includegraphics[width=1.61in]{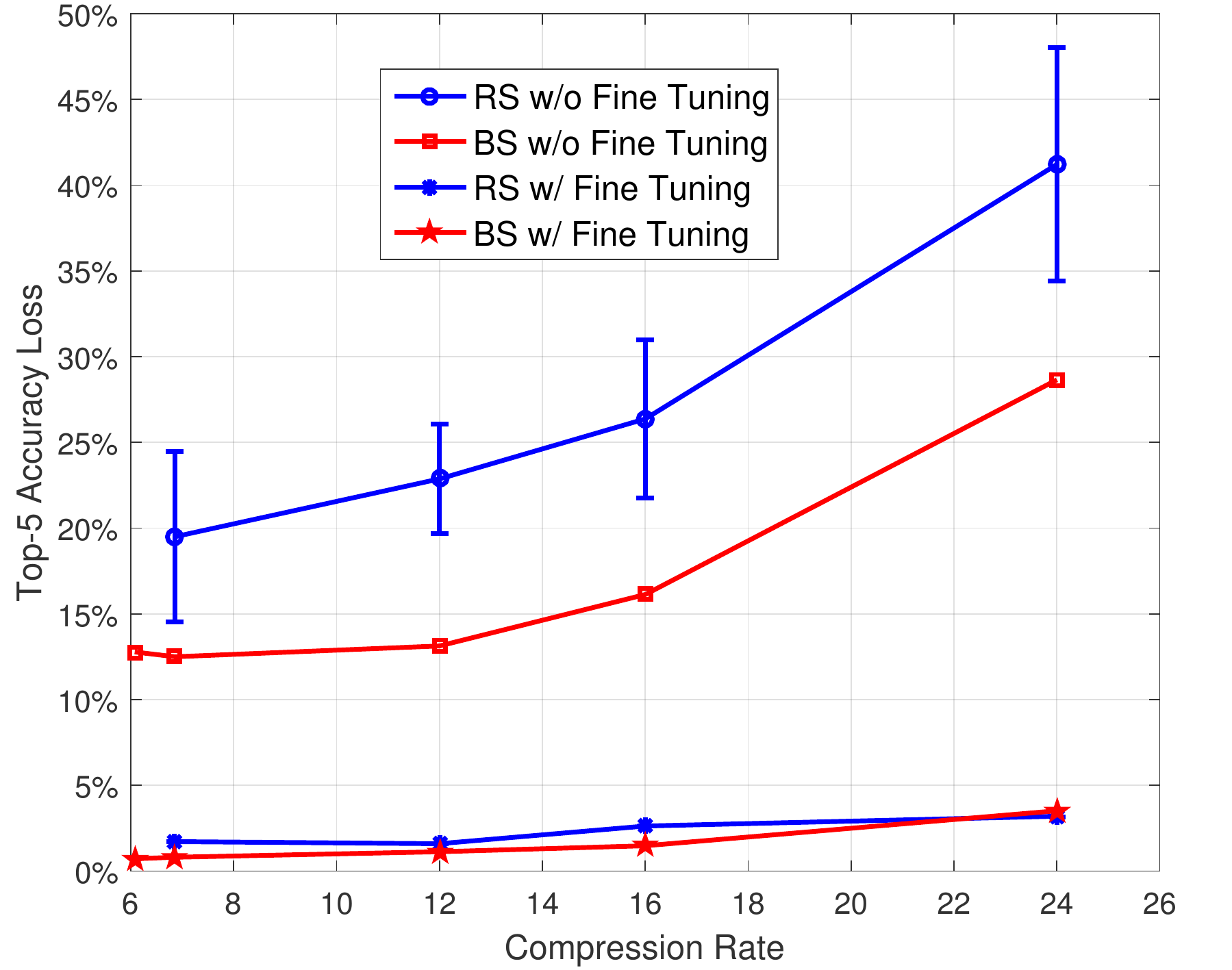} \\
  (a) & (b)
\end{tabular}
\end{center}
\vskip -15pt
\caption{\label{RandomSearchImagenet} Accuracy loss of compressed AlexNet v.s. compression rate. }
\vskip -5pt
\end{figure}

The compression performance is shown in Fig. \ref{RandomSearchImagenet}, and the computational complexity is shown in Table \ref{ImageNetNum}. We can see that with much smaller computational complexity, the proposed backward search can achieve better compression performance than random search. Moreover, the classification performance of proposed scalable compression framework drops little when the compression rate is within 10. The compression performance is comparable to state-of-the-art algorithms at fixed rate, e.g., AlexNet is compressed to 47.6 MB with more than $1\% $ Top-1 accuracy loss in \cite{NIPS2014_5544}.


\section{Conclusions and Future Work}
\label{sec_con}

\begin{table}[tb]
\begin{center}
\begin{tabular}{c|c|c|c|c}
\hline
 Bit Budget (MB) & 35 & 20 & 15 & 10\\
\hline
Compression Rate &  6.86 & 12 & 16 & 24 \\
\hline
BS Number & 25 & 81 & 89 & 126 \\
\hline
RS Number & \multicolumn{4}{c}{150} \\
\hline
\end{tabular}
\caption{\label{ImageNetNum}
Number of configurations tested on ILSVRC12 validation set v.s. compression rate.}
\end{center}
\vskip -10pt
\end{table}

In this paper, we discuss the scalable compression of deep neural networks, and propose a three-stage pipeline: hierarchical quantization of weights, backward search for bit allocation, and fine-tuning. Its efficacy is tested on three different DNNs.
In \cite{ICLR16}, the authors can compress AlexNet to 6.9 MB without loss of accuracy, much smaller than what is achieved in this paper, due to network pruning is used \cite{Pruning}, which removes many small-weight connections from the network. This not only compresses the network, but also reduces the complexity of the implementation. In addition, entropy coding is used in \cite{ICLR16}. However, the quantization in \cite{ICLR16} is fixed and not scalable. This paper focuses on the scalable quantization and adaptive bit allocation. It is also shown from Fig. 7 in \cite{ICLR16} that pruning does not hurt quantization. Therefore the pruning and entropy coding can also be used in our scheme to further improve the performance.
\section*{Acknowledgement}

This work was supported by by NSERC of Canada under grant RGPIN312262, STPGP447223, and RGPAS478109, and NVIDIA University Partnership Program.

\bibliographystyle{abbrv}

%

\end{document}